%% file: main.tex
\documentclass[10pt]{article} 

\usepackage[accepted]{tmlr}

\input{math_commands.tex}

\usepackage{hyperref}
\usepackage{url}

\usepackage{tikz} 
 \usepackage{tikz}
\usetikzlibrary{automata,positioning,calc, arrows}
\usepackage{centernot}
\definecolor{orangeex}{HTML}{FFB366}
\definecolor{blueex}{HTML}{FF9999}
\definecolor{lighbluex}{HTML}{6666FF}
\definecolor{greenex}{HTML}{50A854}
\usepackage{amssymb}
\usepackage{xcolor}
\usepackage{longtable}
\usepackage{booktabs}
\usepackage{subcaption}

\title{Sortability of Time Series Data}

\author{\name Christopher Lohse \email lohsec@tcd.ie \\
      \addr School of Computer Science and Statistics\\
      University of Dublin Trinity College \\
      IBM Research Europe, Dublin\\
      \AND
      \name Jonas Wahl \email jonas.wahl@dfki.de \\
      \addr Deutsches Forschungszentrum für künstliche Intelligenz (DFKI)
      }

\def\month{07}  
\def\year{2025} 
\def\openreview{\url{https://openreview.net/forum?id=OGvmCpcHdV}} 

\begin{document}
  \tikzset{auto shift/.style={auto=right,->,
to path={ let \p1=(\tikztostart),\p2=(\tikztotarget),
\n1={atan2(\y2-\y1,\x2-\x1)},\n2={\n1+180}
in ($(\tikztostart.{\n1})!1mm!270:(\tikztotarget.{\n2})$) -- 
($(\tikztotarget.{\n2})!1mm!90:(\tikztostart.{\n1})$) \tikztonodes}}}

\tikzset{auto shiftdouble/.style={auto=right,<->,
to path={ let \p1=(\tikztostart),\p2=(\tikztotarget),
\n1={atan2(\y2-\y1,\x2-\x1)},\n2={\n1+180}
in ($(\tikztostart.{\n1})!1mm!270:(\tikztotarget.{\n2})$) -- 
($(\tikztotarget.{\n2})!1mm!90:(\tikztostart.{\n1})$) \tikztonodes}}}

\tikzset{>=latex}

\maketitle

\begin{abstract}
 Evaluating the performance of causal discovery algorithms that aim to find causal relationships between time-dependent processes remains a challenging topic. In this paper, we show that certain characteristics of  datasets, such as varsortability \cite{reisach2021beware} and $R^2$-sortability \cite{reisach2023scale}, also occur in datasets for autocorrelated stationary time series. We illustrate this empirically using four types of data: simulated data based on SVAR models and Erdős-Rényi graphs, the data used in the 2019 causality-for-climate challenge \citep{runge2019inferring}, real-world river stream datasets, and real-world data generated by the Causal Chamber of \cite{gamella2024chamber}. To do this, we adapt var- and $R^2$-sortability to time series data. We also investigate the extent to which the performance of continuous score-based causal discovery methods goes hand in hand with high sortability.
 Arguably, our most surprising finding is that the investigated real-world datasets exhibit high varsortability and low $R^2$-sortability indicating that scales may carry a significant amount of causal information.
\end{abstract}

\section{Introduction}\label{sec:intro}


Inferring causal relationships between variables in a multivariate time series setting is an ongoing topic of research in many fields, including economics/econometrics \citep{varian2016causal}, climate science \citep{runge2019inferring, runge2020discovering}, medicine \citep{yazdani2015causal} and neuroscience \citep{bergmann2021inferring}. The process of inferring causal structure from data is often referred to as causal discovery or causal structure learning.
Most of the literature on causal discovery is devoted to the following two types of methods:
    Constraint-based causal structure learning algorithms, such as the PC-algorithm \citep{spirtes1991algorithm} and PCMCI \citep{runge2020discovering} for time series data, use tests of conditional independence to iteratively learn a causal graph.
    Score-based method such as GES \citep{chickering2002optimal} fit a directed acyclic graph (DAG) to the data by optimising a score function, but need to search a large discrete space of DAGs which is an NP-hard problem.

    More recently, \cite{zheng2018dags} have proposed to embed the discrete DAG-search space into a continuous one through a differentiable acyclicity constraint to make the problem amenable to continuous (gradient based) optimisation. The method introduced in \cite{zheng2018dags}, NOTEARS, showed impressive performance on data simulated with linear additive noise models (LANMs). However, as shown by \cite{reisach2021beware} the strong performance of NOTEARS and similar methods on LANM data vanished after the data was normalized. \cite{reisach2021beware} noticed that, before normalization, additive noise model data is highly varsortable, meaning that, on average, the causal order of the system can be recovered well by sorting the variables by the amplitude of their estimated variances. Since high varsortability and good NOTEARS performance were highly correlated, they therefore conjectured that NOTEARS implicitly made use of variance sorting in its optimisation. The issue has recently been revisited by \cite{Ng2024}, who pointed out that (a) NOTEARS does not necessarily perform well in the presence of high varsortability and that (b) normalisation of data generated by LANMs moves the data far away from the assumption of equal noise variances that underlies the NOTEARS methodology. Thus, according to \cite{Ng2024}, NOTEARS should not have been expected to perform well on normalized LANM data in the first place as one of its fundamental assumptions is not satisfied.

Rather than weighing in on this discussion, in this contribution we explore how much var- and $R^2$-sortability (the coefficient of determination which acts as a proxy for the fraction of the variance a variable that is explained by its causal parents introduced in \cite{reisach2023scale}) can be seen in data commonly used in method validation of \emph{time series} causal discovery methods. 
We evaluate the degree of var-/$R^2$-sortability in different datasets as well as the performance of different algorithms for time series causal discovery in the presence/absence of var-/$R^2$-sortability and before/after normalisation. One of the main questions underlying the discussions on sortability is this: how much var-/$R^2$-sortability do we expect to see in real-world data? The answer to this question is likely highly context-dependent, and the question can be notoriously difficult to settle even for a single dataset, since one needs to know the causal ground truth to compute sortability. We investigate sortability scores in two of the rare cases where a causal ground-truth is available, the river flow dataset used in \cite{tran2024estimating} and the Causal Chamber datasets recently published by \cite{gamella2024chamber}. In both cases, we find sortability values quite far from $0.5$ (the case where there is no information in the sortability criterion), meaning that there is information in the variance or $R^2$-values of the data. In the river stream datasets, marginal variances tend to decrease along the causal order (varsortability close to zero) while in the Causal Chamber case, marginal variances increase along the causal order (varsortability close to one). For the $R^2$-score we find the exact opposite: values close to one for some rivers in the river dataset, and values close to zero for some of the causal chamber datasets.

These observations illustrate that discarding scales in causal discovery as arbitrary may be premature, as scales may encode significant causal information. We hypothesise plausible physical explanations for the observed varsortability in the investigated datasets which also call into question the validity of an equal noise variance assumption in these cases. For the river data we know that the the width of the rivers decrease from the source to the mouth which has potentially an influence on the variance of extreme flows.

In more detail, our main contributions are:
\begin{enumerate}
        \item We extend var- and $R^2$-sortability and the simple benchmarking algorithm of \cite{reisach2021beware, reisach2023scale} to the time series setting.
        \item We show empirically that simulated data typically used to evaluate causal discovery algorithms for time series data is varsortable, meaning the amplitude of the marginal variance increases the lower the variable is in the causal ordering of the ground truth summary graph. 
        We also show that, in this case, varsortability is largely driven by contemporanous dependencies.
         \item We demonstrate that there is a positive correlation between the performance of continuous score-based causal discovery algorithms for time series and the varsortability of data generated by structural autoregressive processes.
        \item We investigate sortability of the data used in the 2019 causal discovery challenge\footnote{\url{https://causeme.uv.es/neurips2019/}} by \cite{runge2019inferring} and show that some data sets are highly varsortable and simple benchmark perform well on these. 
        \item We calculate var- and $R^2$-sortability of the river flow dataset used in \cite{tran2024estimating}, and of the recently published datasets generated by the Causal Chamber\citep{gamella2024chamber}. We find that in these real-live datasets, the scale plays an important role for potential causal discovery algorithms.

\end{enumerate}

\section{Preliminaries}

\subsection{Causal discovery for time series}\label{sec:svar}
A \emph{stationary time series graph} (ts-graph) is a directed graph $\mathcal G = (V \times \mathbb Z, \mathcal D), \ V = \{1.\dots,d \}$, whose edges $(i,t-k) \to (j,t)$ are assumed invariant under translation of the time component. In addition, it is typically required that there is a finite maximal lag $\tau_{max} = \max_{i,j \in V} \{ k | ((i,t-k),(j,t)) \in \mathcal D \} < \infty$ and that the contemporaneous component of $\mathcal G$ is acyclic, where $k \in \mathbb{N}$. Any stationary ts-graph induces a directed, potentially cyclic \emph{summary graph } $\mathcal G_{\mathrm{sum}}$ over $V$ that contains a directed edge $(i,j)$ if $(i,t-k) \to (j,t) \in \mathcal D$. The adjacency matrices of $\mathcal{G},\mathcal G_{\mathrm{sum}}$ will be denoted by $\mathbf{W},\mathbf{W}_{\mathrm{sum}}$ respectively.

The aim of most causal discovery methods for time series is to recover either $\mathcal G$ or $\mathcal G_{\mathrm{sum}}$ from observational or interventional data. Often the data is assumed to be generated by a discrete multivariate process $ (\mathbf{X}_t)_{t \in \mathbb Z}, \ \mathbf{X}_t =(X_t^1, ..., X_t^{(d)})$ compatible with $\mathcal{G}$. The process $(\mathbf{X}_t)_{t \in \mathbb Z}$ is typically modelled as a structural vector autoregressive (SVAR) process\citep{hyvarinen2010estimation},  in which case, compatibility with $\mathcal{G}$ means that $(\mathbf{X}_t)_{t \in \mathbb Z}$ follows the evolution rule

\begin{equation}
    X_t^j = \sum_{i \in \mathrm{pa}_{\mathcal{G}_{\mathrm{sum}}}(j)} \sum_{k=0}^{\tau_{max}} a^j_{i,t-k} X_{t-k}^i + \eta^j_t
\end{equation}
for $j \in \{1, ..., d\}$, where $\eta^j$ are Gaussian white noise processes,  $\mathrm{pa}_{\mathcal{G}_{\mathrm{sum}}}(j)$ denotes the parents of node $j$ in the summary graph, and $a^j_{i,t-k}$ is only allowed to be non-zero if $(i,t-k) \to (j,t)$ is an edge in $\mathcal{G}$. SVAR-processes can be considered the time series analogue of additive linear noise models for which sortability was discussed by \cite{reisach2021beware,reisach2023scale}. When generating data from such a model, coefficients are typically randomly drawn, sometimes with a pre-specified proportion of contemporaneous links, and the process is being run until it has converged to a stationary distribution (or is discarded if the distribution is non-stationary). 

\subsection{NOTEARS and Derivatives}
\cite{zheng2018dags} propose the continuous score-based causal discovery method NOTEARS, which embeds the discrete search space of DAGs into a continuous one by using the differentiable function  $h(\mathbf{W}) = tr \: e^{\mathbf{W} \circ  \mathbf{W}} - d$. This function 
is 0 if and only if $\mathbf{W}$ is the adjacency matrix of an acyclic graph and hence measures the \textit{DAGness} of $\mathbf{W}$ \citep{zheng2018dags}. By combining this function with a score evaluating how well the estimated weight matrix $\mathbf{W}$ fits the data, \cite{zheng2018dags} formulate the constrained optimisation problem to find
\begin{equation}
    \min_{\mathbf{W}} \frac{1}{n} ||\mathbf{X} - \mathbf{X}\mathbf{W}||^2_2
\end{equation}

s.t. $\mathbf{W}$ is acyclic, which is modelled by $h$.
$||\cdot||^2_2$ is the Frobenius norm.

The DYNOTEARS algorithm \cite{pamfil2020dynotears} modifies the NOTEARS algorithm to work with time-lagged and auto-correlated dependencies by redefining the optimisation problem to $\min_{\mathbf{W}1 } \ell(\mathbf{W_c}, \mathbf{W_l}) \text{ s.t. } \mathbf{W}_c \text{ is acyclic}$, where $\mathbf{W_l} \in \mathbb{R}^{d \times d \times \tau_{max}}$ is the lagged adjacency matrix and  $\mathbf{W}_c$ is the contemporaneous adjacency of the underlying time series process. As the underlying time series graph to estimate is acyclic if and only if $\mathbf{W}_c$ is acyclic, it suffices to enforce the acyclicity constraint only on $\mathbf{W}_c$ \citep{pamfil2020dynotears}. To ensure sparsity of $\mathbf{W}$, \cite{pamfil2020dynotears} also add an $\ell_1$ penalty term, leading to the constraint optimisation problem 
\begin{equation}
\min_{\mathbf{W}} \frac{1}{2n} ||\mathbf{X} - \mathbf{X}\mathbf{W}_c -\mathbf{X_l}\mathbf{W}_l||^2_2 + \lambda_1 ||\mathbf{W}_c||_1 + \lambda_2 ||\mathbf{W}_l||_1
\end{equation}
s.t. $\mathbf{W}_c$ is acyclic. Here $\lambda_1$ and $\lambda_2$ are two regularisation parameters and $\mathbf{W}_l$ is the lagged adjacency matrix.

The continuous optimisation problems of NOTEARS and DYNOTEARS can be solved efficiently by rewriting the problem using the augmented Lagrangian method and using a numerical solver such as L-BFGS \citep{zheng2018dags, pamfil2020dynotears}.
After applying the numerical optimisation algorithm in both of the algorithms, a threshold $t$ is applied to remove weights close to zero \citep{zheng2018dags,pamfil2020dynotears}.
\subsection{Sortability Criteria}

In the following, we briefly reiterate the two sorting criteria varsortability and $R^2$-sortability introduced by \cite{reisach2021beware} and \cite{reisach2023scale} respectively. 
In essence, both of these approaches calculate a score $s$ of sortability in a comparable manner; $s$ is the measurement of the degree of agreement between the true causal order and the increasing order of a sortability criterion $cri$.

For any causal model containing the variables  $\{X^{(1)}, \ldots, X^{(d)}\}$ with a (non-degenerate)  adjacency matrix $\mathbf{W}$, the sortability score is the fraction of directed paths that start from a node with a strictly lower sortability criterion than the node they end in. Thus the sortability for one selected criterion $cri$ can be calculated as

\small
\begin{equation}
    s := \frac{\sum^{d -1}_{k = 1} \sum_{i \rightarrow j \in \mathbf{W}^k} increasing(cri(X^i), cri(X^j))}{\sum^{d -1}_{k = 1} \sum_{i \rightarrow j \in \mathbf{W}^k} 1} \in [0, 1], \label{eq:sor}
\end{equation}

where

\begin{equation}
increasing(a, b) = \begin{cases} 
      1 & a< b \\
     1/2 & a = b \\
      0 & a >b .
   \end{cases}
\end{equation}
The matrix $\mathbf{W}$ in \autoref{eq:sor} is set to the power $k$, since the $(i,j)$ entry of the $k$-th power of an adjacency matrix of a DAG exactly counts the number of directed paths from $i$ to $j$.
In the case of varsortability \cite{reisach2021beware},
the criterion $cri(X^i) =  \mathrm{Var}(X^i)$ is the marginal variance, whereas in the case of $R^2$-sortability, it is the coefficient of determination $cri(X^i) =R^2(X^i)$ which acts as a proxy for the fraction of the variance of $X_i$ that is explained by its causal parents, (see \cite{reisach2023scale} for details).
Varsortalility $v$ is defined by using the marginal variance as $cri$.
$R^2$-sortability $r$ is defined by using the obtained $R^2$-coefficients as the sorting criterion $cri$ \citep{reisach2021beware, reisach2023scale}.

\cite{reisach2021beware} showed that varsortability is usually high in data generated by LANMs (see the analytical and empirical proof in their paper). 
Based on their sortability criteria, \cite{reisach2021beware} and \cite{reisach2023scale} introduced the baseline methods varsortnregress and $R^2$-sortnregress respectively. These algorithms sort the system variables based on their variances or $R^2$-scores which are estimated by fitting a regression model; these simple benchmark methods are then shown to have similar performance to some state of the art causal discovery algorithms \citep{reisach2021beware, reisach2023scale} on LANM data.

\section{Modified Sortability Criteria for Summary Graphs} \label{sec:sort}


In a time series causal discovery setting, the summary graph $\mathcal{G}_{\mathrm{sum}}$ describing the relationship between the different time-evolving processes may contain cycles. Since the marginal variances $\mathrm{Var}(X^i_t)$ do not depend on the time index $t$ due to the assumed stationarity of the processes, to compute varsortability in this situation, in principle, one could still use \autoref{eq:sor} as is.

However, any pair of processes $X^i$ and $X^j$ that belong to the same strongly connected component of the summary graph (i.e. that are connected by a cycle) would always contribute a $1$ to the numerator and a $2$ to the denominator of \autoref{eq:sor}.


Therefore the presence of cycles would dilute the sortability signal and would naturally push it closer to $1/2$. In other words, $R^2$-, and varsorting of cyclicly connected processes is meaningless, and we are only interested in whether nodes that are not cyclicly connected can be sorted. Our sortability criterion thus becomes 

\begin{equation}
    s := \frac{\sum_{(i,j)\in \mathcal{AP}(\mathcal{G}_{\mathrm{sum}}) }  increasing(cri(X_i), cri(X_j))}{\sum_{(i,j)\in \mathcal{AP}(\mathcal{G}_{\mathrm{sum}}) } 1} \in [0, 1], \label{eq:sor_mod}
\end{equation}
where $\mathcal{AP}(\mathcal{G}') = \{ (i,j) \in V' \times V' \ | \ i \Longrightarrow j, \ j \centernot{\Longrightarrow} i \}$ is the set of admissible node pairs (the long double arrow indicates the existence of a directed path).

\begin{figure}[h!]
    \centering
   
\begin{tikzpicture}[node distance={20mm}, main/.style = {draw, circle}]
\node[main] (1) {\textcolor{black}{$C$}};
\node[main] (2) [above right of=1] {\textcolor{black}{$B$}};
\node[main] (3) [below right of=1] {\textcolor{black}{$D$}};
\node[main] (4) [above right of=3] {\textcolor{black}{$A$}};

\node[right=1mm of 1] {\footnotesize $Cri(C) = 3$};
\node[right=1mm of 2] {\footnotesize $Cri(B) = 2$} ;
\node[right=1mm of 3] {\footnotesize $Cri(D) = 2.5$} ;
\node[right=1mm of 4] {\footnotesize $Cri(A) = 1$};

\draw
(2) edge[auto shift, color=orangeex] node {}  (1)
(1) edge[<->, auto shiftdouble] node {} (2);

\draw
(4) edge[color=blueex, auto shift] node {} (2);
\draw[-, color=orangeex] (4) -- (2);

\draw[->, color=lighbluex, auto shift] (3) -- (1) -- (2);

\draw
(3) edge[color=greenex, auto shift] node {} (1);

\end{tikzpicture}
    \caption{Example of the calculation of $s$ with cycles. }
    \label{fig:example}
\end{figure}

For the example in \autoref{fig:example}, the sortability score $s$ is $\frac{\textcolor{blueex}{1} + \textcolor{orangeex}{1} +\textcolor{greenex}{1} + \textcolor{lighbluex}{0}}{\textcolor{blueex}{1}+\textcolor{orangeex}{1}+\textcolor{greenex}{1} +\textcolor{lighbluex}{1}} = 0.75$.
The contribution of the pairs $(B,C)$ and $(C,B)$ is ignored since the two nodes belong to the black cycle. Cycle-free directed paths that connect admissible node pairs are depicted in colour.
If we would calculate it including cyclicly connected pairs, it would result in $\frac{4}{6} \approx 0.67$.

\paragraph{High varsortability and the performance of DYNOTEARS}

Based on heuristics and empirical findings, \cite{reisach2021beware} argue that high variability causes gradient-based optimisation algorithms such as NOTEARS\citep{zheng2018dags} to favour graphs whose edges point in the direction of increasing marginal variances during their first optimisation step. 
Since DYNOTEARS is a time-series adaptation of NOTEARS, that essentially coincides with NOTEARS on the contemporaneous components of a ts-graph, \cite{reisach2021beware}'s arguments might be applicable to (the contemporaneous part of) DYNOTEARS as well.

However, in the recent work \cite{Ng2024} provide examples in which continuous optimization-based methods can not perform well in the presence of high varsortability. \cite{Ng2024} also give an alternative explanation for why continuous structure learning performs significantly worse after standardisation of LANM data. Continuous structure learning assumes equal noise variances for all variables in the system, which is violated in the standardised model leading to poor performance.

\section{Sortnregress for Time Series Graphs}\label{sec:sortregress}

In order to have simple algorithms that exploit high $R^2$- and varsortability, we present our time series adapted sortnregress algorithm based on sortnregress from \cite{reisach2021beware, reisach2023scale}. To estimate contemporaneous dependencies, we use the standard sortnregress algorithm, which consists of two steps:
\begin{enumerate}
    \item Sort nodes by increasing marginal variance or $R^2$-score.
    \item  Each node is regressed on its predecessor, determined by order, using a penalised regression technique. As described in \cite{reisach2021beware}, LASSO regression is used, using the Bayesian Information Criterion (BIC) for hyperparameter selection.
\end{enumerate}
This gives us an estimated contemporaneous adjacency matrix $\hat{\mathbf{W}}_c$.
A random sortnregress algorithm is also used, where we determine the order of the variables randomly using i.i.d. Bernoulli trials.
To estimate lagged dependencies between variables, we use the same first step and change the second step:
We now regress each node on each of its predecessors $p_{i, t}$, where $t \in [1, \tau_{max}]$ indicates the time lag. After this step we have an estimated lagged adjacency matrix $\hat{\mathbf{W}}_l$.

\section{Numerical Experiments \& Results }
In the following section, we first give an overview of the evaluation metrics which are used for all the considered datasets. After that we outline the setup and the results for each dataset. 
We conduct our experiments in Python, using the TIGRAMITE library\footnote{\url{https://github.com/jakobrunge/tigramite}} for simulating data with SVAR models.
When assessing the performance of different algorithms across a range of sortability values, the  hyperparameters of the DYNOTEARS algorithm are set to $\lambda_1 = \lambda_2 = 0.05$. The weight threshold is set to $0.1$. As a constraint-based comparison algorithm, PCMCI$^+$\citep{runge2020discovering} is run with $\alpha = 0.01$ and the ParCorr conditional independence test. We further use the varsortnregress, $R^2$-sortnregress and random regress algorithms as described in Section \ref{sec:sortregress}. 

We assess the \( F1 \)-score using the formula:
\[
   F1 =  \frac{TP}{TP + 0.5(FP+FN)}
\]
to gauge the performance of the selected algorithms concerning the comparison between the estimated binary time series adjacency matrices \( \hat{\mathbf{W}} \) and the ground truth \( \mathbf{W} \).

Here, \( TP \) represents the number of true positives, \( FP \) represents the number of false positives, and \( FN \) represents the number of false negatives for edge detection.


We refer to this metric as the overall $F1$-score. Additionally, we calculate the $F1$-scores comparing the estimated contemporaneous adjacency matrix $\hat{\mathbf{W}_c}$ with the true contemporaneous adjacency matrix $\mathbf{W}_c$, and the $F1$-scores comparing the estimated lagged adjacency matrix $\hat{\mathbf{W}_l}$ with $\mathbf{W}_l$. These metrics are denoted as $F1$-contemp and $F1$-lagged, respectively. In cases where only information about the summary graph is available, we calculate the $F1$-score between $\mathbf{W}_{sum}$ and the estimated summary adjacency matrix $\hat{\mathbf{W}}_{sum}$.

\subsection{NEURIPS Competition Data}
\paragraph{Setup}

We also assess var- and $R^2$-sortability on the 2019 Causality-for-Climate-competition\citep{runge2019inferring} data. This dataset is relevant not only because it was used in the competition, but also because it follows the same structure as the CauseMe platform \citep{munoz2020causeme}, which is widely used to evaluate causal discovery algorithms \citep{bussmann2021neural, runge2020discovering}. The dataset includes simulated and partially simulated datasets of varying complexity, including high-dimensional datasets and non-linear dependencies, with 100, 150, 600 or 1000 realisations for each dataset specification.  We excluded the datasets with missing values.

We set the maximal time-lag in our methods $\tau_{max}$ following the description of the respective dataset (ranging from 3 to 5).

\paragraph{Results}

As illustrated in \autoref{tab:neurips}, we observe a varsortability above $0.5$ for all data except the logistic model, which has a varsortability of $0$, meaning that each causal child has a lower marginal variance than its parent. In particular, the realistic climate and weather models have a high varsortability, with a mean of over $0.86$ for all of them. For $R^2$-sortability, we observe a value around $0.5$ for most of the realistic models, with the exception of the FinalCLIM models with values of $0.25$ and $0.16$ for the 5 and 40 variable datasets respectively. The linear and logistic models have scores between $0.5$ and $0.6$. For the data sets with high varsortability (var $>0.8$ varsortnregress outperforms or is en par with PCMCI$^+$.

\begin{table}[h!]
\caption{Mean Sortability criteria and F1-scores for different benchmark algorithms over different realisations on the NeurIps competition data. We set $\tau_{max} = 3$.}
\label{tab:neurips}
\resizebox{\textwidth}{!}{%
\begin{tabular}{@{}lllllll@{}}
\toprule
Dataset                  &           var & $R^2$ &         PCMCI$^+$ & varsortnregress  & $R^2$-sortnregress & random         \\ \midrule
Testlinear-VAR\_N-10\_T-150               & $0.59\pm0.17$  & $0.61\pm0.19$    & - & -  & -& -\\
Testlinear-VAR\_N-100\_T-150                 & $0.65\pm0.11$  & $0.64\pm0.12$  & -               & -                & -               & -               \\
Testnonlinear-VAR\_N-20\_T-600             & $0.56\pm0.13$  & $0.57\pm0.14$  & $0.25 \pm 0.07$ & $0.10 \pm 0.07$  & $0.10 \pm 0.06$ & $0.15 \pm 0.06$ \\
Finallinear-VAR\_N-10\_T-150               & $0.62\pm0.18$  & $0.62\pm0.18$   & $0.15 \pm 0.10$ & $0.17 \pm 0.09$  & $0.16 \pm 0.09$ & $0.20 \pm 0.09$ \\
Finallinear-VAR\_N-100\_T-150              & $0.66\pm0.11$  & $0.63\pm0.11$  & -               & -                & -               & -               \\
FinalCLIM\_N-5\_T-100                       & $0.91\pm0.13$  & $0.29\pm0.25$   & $0.40 \pm 0.13$ & $0.51 \pm 0.23$  & $0.23 \pm 0.18$ & $0.31 \pm 0.19$ \\
FinalCLIM\_N-40\_T-100                      & $0.9\pm0.09$   & $0.19\pm0.11$  & -               & -                & -               & -               \\
FinalCLIMnoise\_N-5\_T-100                   & $0.91\pm0.13$  & $0.3\pm0.25$ & $0.32 \pm 0.16$ & $0.43 \pm 0.22$  & $0.21 \pm 0.18$ & $0.28 \pm 0.21$ \\
FinalCLIMnoise\_N-40\_T-100                 & $0.9\pm0.09$   & $0.23\pm0.13$  & -               & -                & -               & -               \\
Finallogistic-largenoise\_N-5\_T-150\_medium & $0.21\pm0.32$  & $0.58\pm0.35$ & $0.42 \pm 0.23$ & $0.02 \pm 0.10$  & $0.04 \pm 0.13$ & $0.04 \pm 0.13$ \\
FinalWEATHnoise\_N-5\_T-1000                 & $0.77\pm0.21$  & $0.5\pm0.3$   & $0.32 \pm 0.15$ & $0.25 \pm 0.20$  & $0.20 \pm 0.20$ & $0.26 \pm 0.18$ \\
FinalWEATHnoise\_N-10\_T-1000                & $0.27 \pm 0.11$ & $0.23 \pm 0.15$  & $0.17 \pm 0.12$ & $0.21 \pm 0.13$ \\
FinalWEATH\_N-10\_T-1000                    & $0.84\pm0.16$  & $0.52\pm0.22$  & $0.34 \pm 0.10$ & $0.32 \pm 0.14$  & $0.22 \pm 0.12$ & $0.27 \pm 0.12$ \\
FinalWEATH\_N-5\_T-1000                      & $0.81\pm0.2$   & $0.47\pm0.31$& $0.36 \pm 0.14$ & $0.34 \pm 0.18$  & $0.22 \pm 0.17$ & $0.31 \pm 0.17$ \\ \bottomrule
\end{tabular}%
}
\end{table}



\subsection{Data Generation with Erdős–Rényi Graphs and SVAR Models}

\paragraph{Setup}
In order to investigate var- and $R^2$-sortability for datasets used to evaluate continuous score-based causal discovery methods, we replicate one of the two data generation methods used by \cite{pamfil2020dynotears}.

Following \cite{pamfil2020dynotears, zheng2018dags}, when generating random time series graphs, we use 
Erdős–Rényi Graphs (ER Graphs) \citep{newman2018networks} to draw the contemporaneous edges with i.i.d Bernoulli trials. Sampling only lower triangle entries of the contemporaneous adjacency matrix and then permuting the node order ensures that the contemporaneous adjacency matrix $\mathbf{W}_c$ is acyclic. \cite{pamfil2020dynotears} sample the graph to ensure a pre-specified mean degree $d_c$ for the contemporaneous dimension and $d_l$ for the lagged dependencies for a total number of variables $d$. 

In order to ensure that the expected mean degree is $d_c$ the probability of each Bernoulli trial is set to $d_c/(d-1)$, where $d$. The edge coefficients are sampled uniformly at random from $[-2, -0.5]  \cup [0.5, 2.0]$ \citep{pamfil2020dynotears}. Again following \cite{pamfil2020dynotears}, the edge weights for lagged variables are sampled depending on $t$ from $[-0.5 \alpha, -0.3\alpha] \cup [0.3 \alpha, 0.5 \alpha]$, where $\alpha = 1/\delta^{t-1}$. The weight decay $\delta >1$ reduces the influence (weights) of variables further back in time.
We  randomly sample Erdős-Rényi graphs with degree $d_c = 4$ for the contemporaneous dimension, and for each lag we set $d_l= 1$; $\delta$ is set to $1.1$.  

We  examine sortability values for different numbers of nodes $d \in \{10, 20, 50, 100\}$. For each number of nodes we randomly generate $500$ different graphs and generate $n = 500$ samples per graph. The samples are taken after a burn-in period to ensure stationarity. 
We then calculate the overall varsortability, the varsortability of only the contemporanous dependencies and the varsortability of all all the lagged dependencies.

We also want to investigate whether varsortability is driven by contemporaneous or lagged dependencies. This is why we compute var- and $R^2$-sortability over a grid of $d_c, d_l \in [0, 0.5, 1, 2, 3, 4, 6, 8]$. We do this for $d = 10$ and $d= 20$ nodes.

We further investigate the influence of the two sortability criteria on the performance of score and constraint-based algorithms. In order to do so, we generate data for $d = 10$ variables, which has varying sortability values. We then randomly draw $m = 30$ samples per sortability interval, which we set to $[0, 0.2], [0.2, 0.4], [0.4, 0.6], [0.6, 0.8], [0.8, 1]$. 
We report the performance of the following algorithms for varying var- and $R^2$-sortability: DYNOTEARS, DYNOTEARS standardised (run after standardising the data), PCMCI$^+$, varsortnregress/$R^2$-sortnregress and randomregress. This means that for DYNOTEARS, the data has a varsortability as defined by the respective bin before standardising (after standardisation the varsortability is always $0.5$).

\paragraph{Results}
We observe that the overall mean varsortability for the ER-SVAR data used by \cite{pamfil2020dynotears} ranges from $0.58$ for ten nodes to $0.54$ for 100 nodes. We do not see a trend in varsortability for different numbers of nodes. The varsortability of the contemporaneous dimension is around $0.7$ for all numbers of nodes. The lagged varsortability is around $0.54$ for all numbers of nodes except 10 nodes where it is $0.56$. The $R^2$ sortability is around $0.5$ for all numbers of nodes. Varsortability of the contemporaneous component of $\mathcal{G}$ is always above $0.7$ and higher than the overall varsortability. Consequently, lagged varsortability is always lower than contemporanous and overall varsortability.
The detailed results can be found in \autoref{app:ER}.

\begin{figure}[h!]
    \centering
    \begin{subfigure}{0.48\textwidth}
        \centering
        \includegraphics[width=\textwidth]{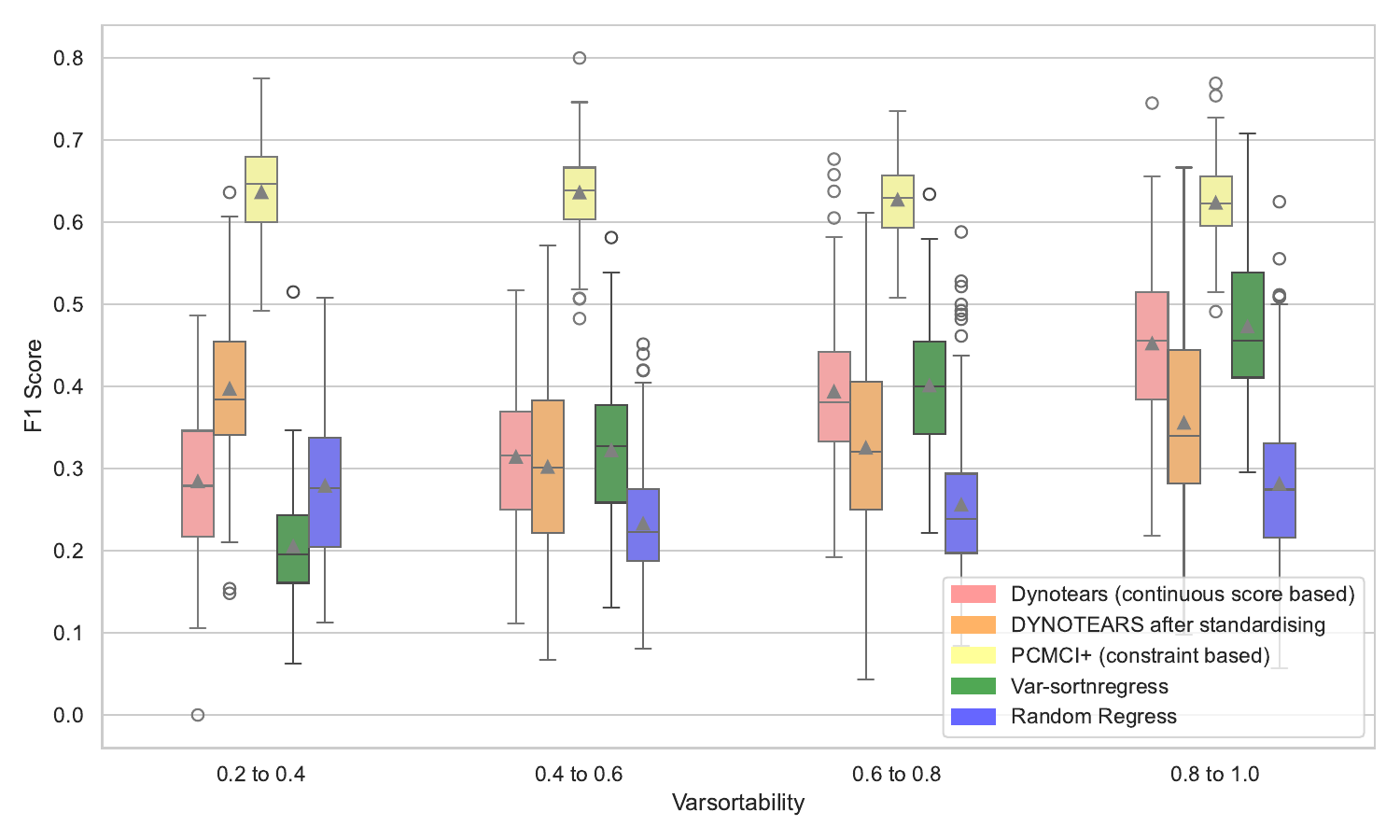}
        \caption{Performance of different algorithms for varying varsortability.}
        \label{fig:varsorts_bench}
    \end{subfigure}
    \hfill
    \begin{subfigure}{0.48\textwidth}
        \centering
        \includegraphics[width=\textwidth]{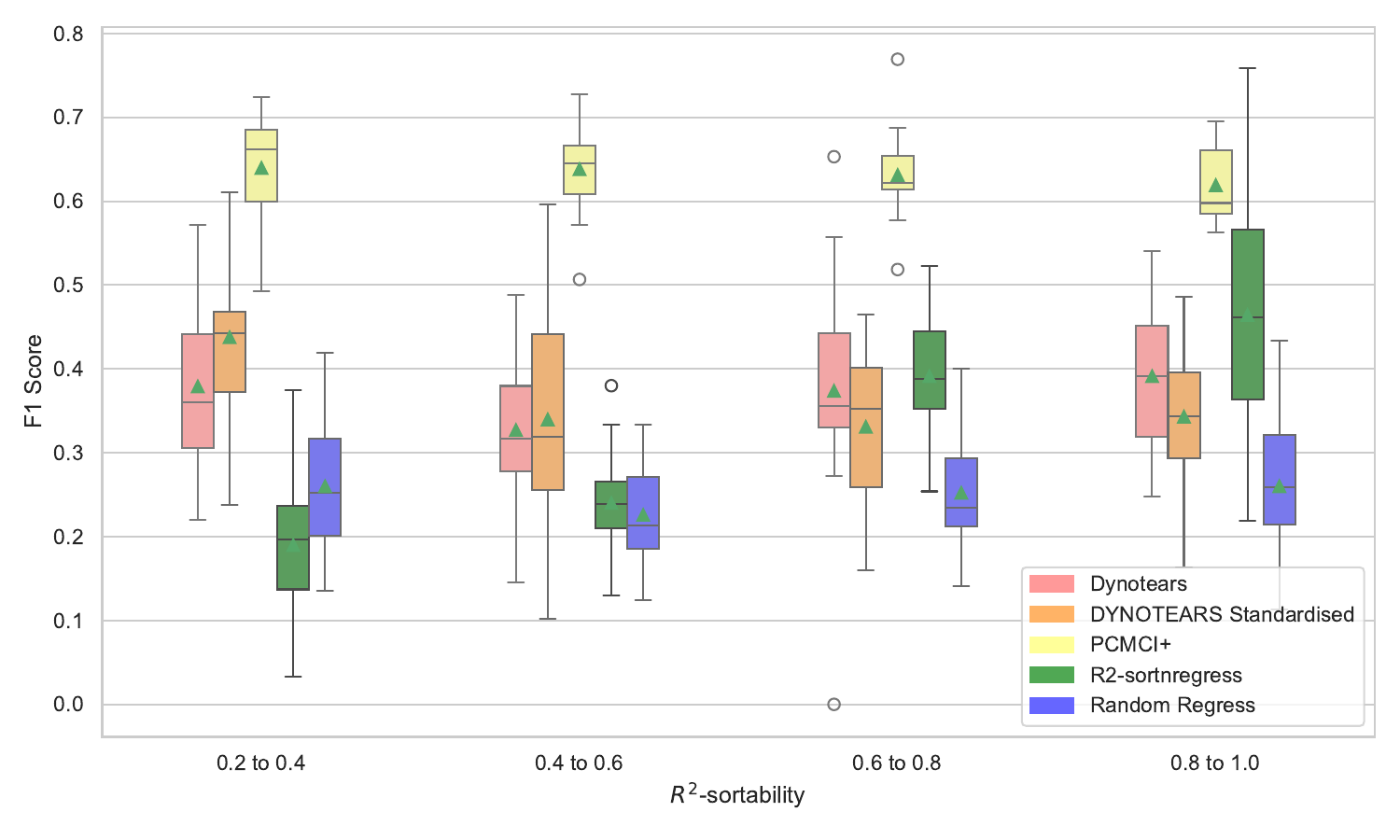}
        \caption{Performance of different algorithms for varying $R^2$-sortability.}
        \label{fig:r2sorts_bench}
    \end{subfigure}
    \caption{Comparison of algorithm performance under different sortability conditions: (a) Varying varsortability, (b) Varying $R^2$-sortability.}
    \label{fig:sorts_bench}
\end{figure}

\autoref{fig:varsorts_bench} shows that the higher the varsortability, the higher the F1-score of DYNOTEARS. The varsortnregress benchmark model also improves with higher varsortability and seems to outperform the DYNOTEARS algorithm for varsortability values higher than 0.6. The constraint-based PCMI$^+$ algorithm does not seem to be as affected by varsortability as the randomregress algorithm. The DYNOTEARS algorithms perform better on standardised data for low varsortability values before standardisation.

For high $R^2$-sortability values, we observe a different behaviour: Both DYNOTEARS and PCMCI$^+$ seem unaffected by varying values. Again the $R^2$-sortnregress algorithm seems to outperform DYNOTEARS for $R^2$-sortability values of $0.6$ or higher (see \autoref{fig:r2sorts_bench}).

\begin{figure}[h!]
    \centering
    \includegraphics[width = 0.48 \textwidth]{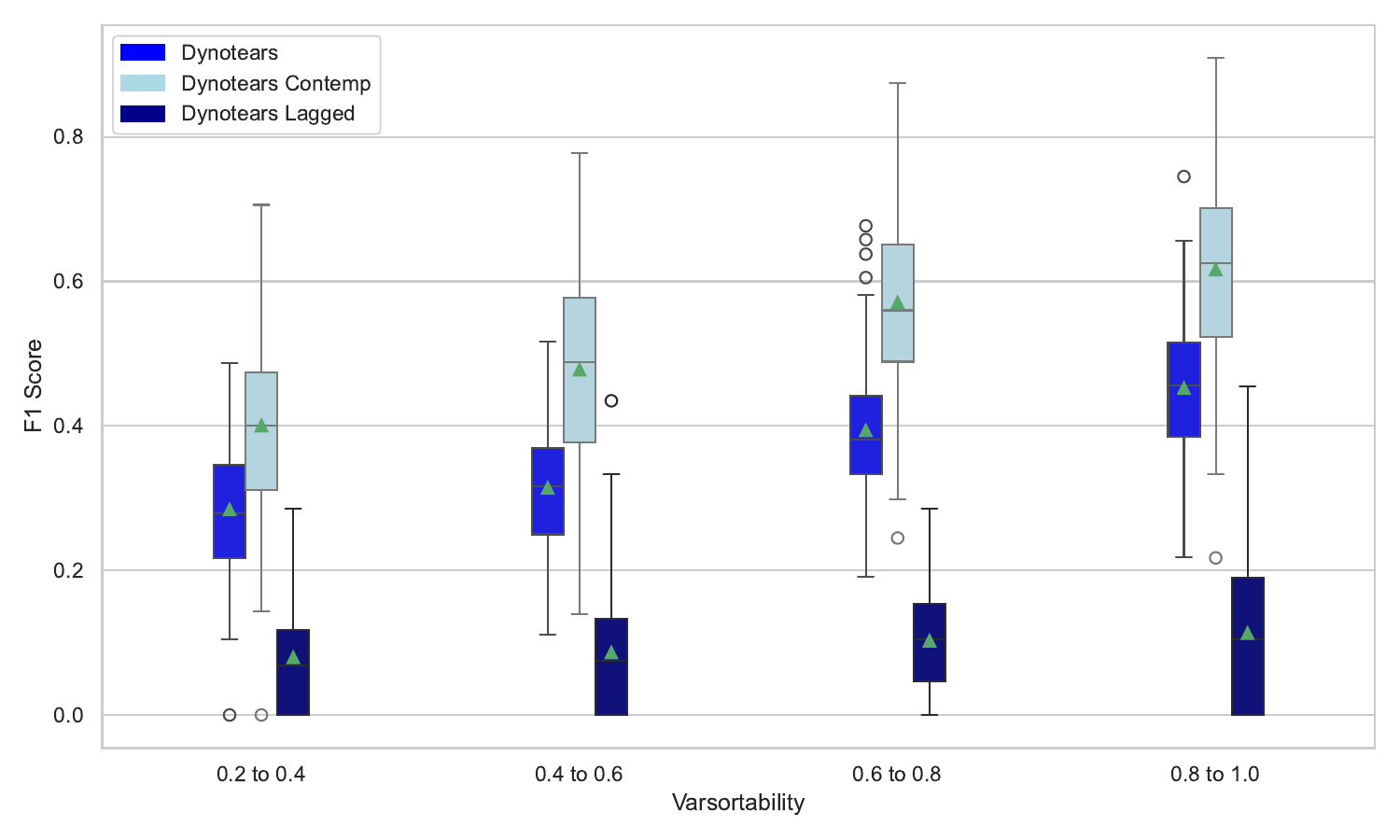}
    \caption{F1-score of Contemporaneous and lagged Dependencies DYNOTEARS for different varsortability values.}
    \label{fig:varsorts_benchdyno}
\end{figure}

We also investigate whether high varsortability leads to a higher F1-score of DYNOTEARS due to better identification of contemporaneous edges or lagged dependencies. As we can see in \autoref{fig:varsorts_benchdyno}, the effect of increasing F1-scores with increasing varsortability is even higher for contemporaneous dependencies. The F1-score for lagged dependencies seems to be unaffected. Moreover, we observe in our experiments that the contemporaneous F1-score has a Pearson correlation of more than $0.6$ with the varsortability score. Furthermore, higher weighting thresholds of the DYNOTEARS algorithm seem to increase the influence of varsortability on the performance of DYNOTEARS as measured by the F1-score. 

We also want to determine if different degrees of contemporaneous and lagged dependencies influence the overall varsortability of the generated data. In general we observed that a higher value for $d_c$ increases the varsortability and higher values of $d_l$ seem to decrease it. If there are more nodes the varsortability seems to be higher. If there are no contemporaneous edges we observe varsortability values of around $0.5$.
For $R^2$-sortability, the  mean score is not affected by the degree or number of nodes. The detailed results can be found in \autoref{app:degree}.

\subsection{Extremal River Flow Problem}
\paragraph{Setup}
Next, we investigate the two sortability criteria on four real-world datasets modelling extremal river flow previously used in \citep{tran2024estimating}. The aim of the problem is to recover the direction and connections of a river network with only extreme flow measurements at certain stations, without knowing the location of these stations. This is a time-dependent process, as an extreme measurement recorded at one station at time $t-1$ may lead to an extreme measurement at another station at time $t$ \citep{tran2024estimating}. Having the ground truth river network allows us to report $R^2$ and varsortability on this real-world dataset and investigate whether high values can also occur on real time series data. We also include the $F1$-scores of selected benchmark algorithms for a $\tau_{max}$ value of 3 as this gave the best results. Following \cite{tran2024estimating}, we treat the downstream river direction as the causal ground truth $\mathcal{G}_{sum}$, which perhaps may be controversial due to the fact that downstream extreme events might have damming effects further upstream. Nevertheless, even if one is not comfortable calling the flow direction causal, the results below show that varsorting is able to cover the 'flow graph'. 

As we only know the ground truth/flow graph on the summary level, i.e. we know $\mathcal{G}_{sum}$, we calculate the $F1$-score between the ground truth for $\mathbf{W}_{sum}$ and the estimated adjacency matrix $\hat{\mathbf{W}}_{sum}$ for the following algorithms: 
varsortnregress, $R^2$-sortnregress, randomregress, PCMCI$^+$ and reversed varsortnregress. We use reversed varsortnregress as we observed very low varsortability values and wanted to investigate if an algorithm that orders by decreasing variance could exploit this fact.

\paragraph{Results}
As illustrated in \autoref{tab:river}, varsortability for the entire river network is below 0.5, with a varsortability value for the Danube close to 0. This indicates that as the river network follows a tree structure, the variance of the extremal river velocities for nodes decreases with increasing distance from the river source. In other words, the real-life dynamics of the river flow systems entail sortable marginal variances.
It can be observed that for high $R^2$-sortability, the $R^2$-sortnregress algorithm performs as well as PCMCI$^+$. Conversely, for low varsortability values, the reversed varsortability algorithm, while being superior to random, is unable to match the performance of PCMCI$^+$.
Notably, the varsortnregress algorithm outperforms the reverse varsortnregress algorithm on the upper Colorado dataset even though the varsortability is below 0.5.
\begin{table}[h!]
\caption{Varsortability(var) and $R^2$-sortability ($R^2$)and $F1$-scores of different algorithm reported on the extremal river flow datasets. We set $\tau_{max} = 3$ as this leads to the best results.} \label{tab:river}
\centering
\resizebox{\columnwidth}{!}{%

\begin{tabular}{@{}lccccccc@{}}
\toprule
Dataset         & var & $R^2$ & PCMCI$^+$ & varsortnregr. & varsortnregr. rev. & $R^2$-sortnregr. & randomregr. \\ \midrule
danube          & 0.07 & 0.82 & 0.56      & 0.08          & 0.20               & 0.18             & 0.11         \\
lower-colorado  & 0.34 & 0.61 & 0.36      & 0.08          & 0.08               & 0.08             & 0.10         \\
middle-colorado & 0.29 & 0.94 & 0.26      & 0.09          & 0.23               & 0.26             & 0.22         \\
upper-colorado  & 0.43 & 0.95 & 0.41      & 0.26          & 0.11               & 0.30             & 0.22         \\ \bottomrule
\end{tabular}
}

\end{table}
We also tried to run DYNOTEARS on this dataset. However, the algorithm did not converge after a couple of hours run time as it did not manage to satisfy the acyclity constraint, at least for our choices of hyperparameters.
\subsection{Causal Chamber Data}
\paragraph{Setup}
We now investigate the values of the two sortability criteria for data generated by the recently introduced Causal Chamber \citep{gamella2024chamber}, which provides a toolbox consisting of real physical systems that can be used to evaluate causal discovery or other AI algorithms on real data. 

We use each dataset contained in their PYTHON library  and calculate the var- and $R^2$-sortability for each dataset and each experiment in the dataset\footnote{\url{https://github.com/juangamella/causal-chamber}}. We then calculate the mean and standard deviation of the different datasets, where one value is one experiment performed on the dataset.
We again run benchmarks  on varsortnregress, $R^2$-sortnregress, randomregress, PCMCI$^+$ and evaluate the $F1$-score between $\mathbf{W}_{sum}$ and $\hat{\mathbf{W}}_{sum}$.

\begin{table}[h!]
\centering
\caption{Mean and standard deviation of var- and $R^2$-sortability as well as F1-score of benchmark algorithms obtained on each dataset with different experiments. The standard deviation is given after $\pm$. We set $\tau_{max} = 2$.}\label{tab:reschamber}
\resizebox{1 \columnwidth}{!}{%
\begin{tabular}{@{}lllllll@{}}
\toprule
Dataset                  &           var & $R^2$ & PCMCI$^+$ & varsortnregress & $R^2$-sortnregress & random \\ \midrule
lt\_camera\_test\_v1            & 0.94 $\pm$ 0.01 & 0.25 $\pm$ 0.24 & 0.05 & 0.30 & 0.17 & 0.14\\
lt\_camera\_validate\_v1        & 0.99 $\pm$ 0.02 & 0.01 $\pm$ 0.03 & 0.08 & 0.37 & 0.05 & 0.25\\
lt\_camera\_walks\_v1           & 0.95 $\pm$ 0.0  & 0.23 $\pm$ 0.07 & 0.31 & 0.34 & 0.13 & 0.22\\
lt\_color\_regression\_v1       & 0.94 $\pm$ 0.02 & 0.15 $\pm$ 0.06 & 0.19 & 0.25 & 0.18 & 0.19\\
lt\_interventions\_standard\_v1 & 0.94 $\pm$ 0.03 & 0.46 $\pm$ 0.04 & 0.26 & 0.18 & 0.12 & 0.08\\
lt\_malus\_v1                   & 0.98 $\pm$ 0.02 & 0.02 $\pm$ 0.01 & 0.20 & 0.24 & 0.25 & 0.19\\
lt\_test\_v1                    & 0.96 $\pm$ 0.03 & 0.01 $\pm$ 0.02 & -    & -    & -    & -\\
lt\_validate\_v1                & 0.99 $\pm$ 0.02 & 0.02 $\pm$ 0.02 & 0.34 & 0.30 & 0.29 & 0.13\\
lt\_walks\_v1                   & 0.9  $\pm$ 0.08 & 0.24 $\pm$ 0.04 & 0.16 & 0.38 & 0.22 & 0.33\\
wt\_bernoulli\_v1              & 0.97 $\pm$ 0.06 & 0.06 $\pm$ 0.05 & -    & -    & -    & -\\
wt\_changepoints\_v1           & 0.94 $\pm$ 0.01 & 0.14 $\pm$ 0.02 & 0.12 & 0.14 & 0.09 & 0.06\\
wt\_intake\_impulse\_v1        & 1.00 $\pm$ 0.00 & 0.32 $\pm$ 0.08 & -    & -    & -    & -\\
wt\_pc\_validate\_v1           & 0.78 $\pm$ 0.00 & 0.14 $\pm$ 0.0  & -    & -    & -    & -\\
wt\_pressure\_control\_v1      & 0.92 $\pm$ 0.0  & 0.36 $\pm$ 0.0  & 0.21 & 0.29 & 0.26 & 0.25\\
wt\_test\_v1                   & 0.94 $\pm$ 0.08 & 0.14 $\pm$ 0.13 & -    & -    & -    & -\\
wt\_validate\_v1               & 0.97 $\pm$ 0.05 & 0.03 $\pm$ 0.04 & -    & -    & -    & -\\
wt\_walks\_v1                  & 0.92 $\pm$ 0.03 & 0.28 $\pm$ 0.06 & -    & -    & -    & -\\ \bottomrule
\end{tabular}
}
\end{table}

\paragraph{Results}
We observe very high varsortability values for all datasets;  almost all datasets have values over $0.9$ with the wt\_pc\_validate\_v1  dataset beeing the only exception at $0.78$ as shown in \autoref{tab:reschamber}.
All $R^2$-sortability values are below $0.25$. Most of them are between $0.2$ and $0.32$.
The lt\_camera\_validate\_v1, lt\_malus\_v1, lt\_test\_v1, wt\_bernoulli\_v1 and wt\_validate\_v1 have values very close to $0$. The values for each individual experiment contained in one dataset can be found in \autoref{app:chamber}.
As shown in Table \autoref{tab:reschamber} varsortnregress outperforms other causal discovery algorithms on 9 out of 10 of the selected Causal Chamber datasets. We only report F1-scores for 10 of the data sets since for  For wt\_pc\_validate\_v1 there are not enough samples to select the hyperparameter for sortnregress and the other 6 data sets have over $50,000$ samples resulting in very long execution times.






\section{Discussion and Conclusion}

In general, both varsortability and $R^2$-sortability are present in both simulated and real datasets for benchmarking causal discovery algorithms. 
In line with \cite{reisach2021beware}, we observe that DYNOTEARS, which is a NOTEARS based algorithm, seems to perform better when varsortability is higher, which is in line with our hypothesis in \autoref{sec:sort}.

Furthermore, we observe that the data used in the NEURIPS competition\citep{runge2019inferring} are highly varsortable, especially the realistic datasets, by our defined time series varsortability metric. This is also reflected by the good performance of sortnregress, which should be too simple to perform as good as the more sophisticated PCMCI$^+$ algorithm. 

The low varsortability in the simulated var models could potentially be due to the fact that these datasets do not have contemporaneous dependencies, and as we showed earlier, contemporaneous dependencies seem to be the driver for high varsortability in simulated time series data. This is hardly surprising given that a team exploiting this effect won the competition\citep{weichwald2020causal}. 
As for $R^2$-sortability, we see that high values can lead to better performance of simple benchmark algorithms. However, $R^2$-sortability values observed on the NEURIPS and ER Datasets seem to be too low to be exploited by the sorting algorithm.

The low varsortability value observed for the river data, particularly for the Danube, may be attributed to the fact that the width and catchment area of a river increase from the source to the mouth, resulting in a reduction in the impact of extreme flows on river velocity closer to the river's mouth. Consequently, the marginal variance decreases. This serves to illustrate the importance of scales in real dataset.
The data for the other rivers only covers parts of the river, which probably results in a less intense effect and higher values for varsortability.

In addition, we can see that the Causal Chamber data is highly varsortable overall while having low $R^2$-values. This could be due to the fact that the variables controlled by the user are high in the causal order. We conjecture that deeper in the system and further from the user-controlled variables, dynamic turbulence and other sources of noise start having a larger and larger influence.  Thus, unexplained noise is higher the lower we are in the causal order. This is in line with  the low $R^2$-sortability values as the causal parents explain less and less down the causal order. The increase in noise variance also drives the increase in total marginal variance and affects varsortability in this way.

\paragraph{Limitations}
This paper is an empirical study and we do not determine analytically why high varsortability leads to better performance of score based causal discovery algorithms for time series data. While we believe that our explanations for varsortability in the considered real-world datasets are plausible, they should be treated cautiously as hypotheses only. The only thing that we can say with certainty that sortability is highly context-dependent and therefore discarding scales as arbitrary for causal discovery seems premature.

\paragraph{Conclusion}
In conclusion, our paper represents an empirical extension of the work of \cite{reisach2021beware, reisach2023scale}. We demonstrate that high var-sortability occurs in SVAR-simulated time series data, resulting in enhanced performance of continuous score-based causal discovery algorithms assuming equal noise variance. Moreover, in some settings our simple benchmark algorithms outperformed or were en par with more sophisticated algorithms in the presence of high sortability.
Consequently, we advise caution when assuming equal noise variance for time series causal discovery algorithms. Furthermore, it may be advisable to examine simulated data for high varsortability before using them as benchmark data, as was done in the 2019 NEURIPS competition\citep{runge2019inferring}. Finally, we observe high and low varsortability, as well as $R^2$-sortability, in two different types of real-life datasets: the Causal Chamber data is generated in a controlled environment while the river flow dataset is measured in an uncontrolled environment. This indicates that var- and $R^2$-sortability in auto-correlated time series data is not solely a phenomenon observed in simulated data.
For this reason, we believe that marginal variances may contain relevant causal information and exploiting variance or inverse variance sorting may be justified in some situations, if one can combine it with physical reasons for one or the other (even though these physical reasons might already eliminate the need for causal discovery in the first place). Furthermore, the observed $R^2$-sortability scores indicate that the assumption of equal noise variance is equally tricky as the relative fraction of unexplained variance may change throughout the graph and unequal noise variances are one possible reason for this.


\subsubsection*{Acknowledgments}
JW was supported by the European Research Council (ERC) Starting Grant CausalEarth under the European Union’s Horizon 2020 research and innovation program (Grant AgreementNo. 948112, PI Prof. Jakob Runge) as well as by the European Regional Development Fund (ERDF) and the German Federal State of Saarland as part of the project (To)CERTAIN. During the initial phase of this project, JW was employed at the Technical University of Berlin and affiliated with the DLR Institute of Data Science as a guest researcher.

\bibliography{main}
\bibliographystyle{tmlr}
\newpage

\appendix

\section{Var and R2-sortability for different sizes of ER Graphs}\label{app:ER}

\begin{table}[h!]
\caption{Mean of and standard deviation denoted by $\pm$ of varsortability(var) and $R^2$-sortability ($R^2$) for ER Graphs with a different number of nodes. For each number of nodes $500$ random graphs have been created and $500$ samples have been generated.}\label{tab:ER}
\resizebox{\columnwidth}{!}{%
\begin{tabular}{@{}lllllllll@{}}
\toprule
        & \multicolumn{2}{c}{10 Nodes}                        & \multicolumn{2}{c}{20 Nodes}                        & \multicolumn{2}{c}{50 Nodes}                        & \multicolumn{2}{c}{100 Nodes}                       \\ \cmidrule(l){2-9} 
        & \multicolumn{1}{c}{var} & \multicolumn{1}{c}{$R^2$} & \multicolumn{1}{c}{var} & \multicolumn{1}{c}{$R^2$} & \multicolumn{1}{c}{var} & \multicolumn{1}{c}{$R^2$} & \multicolumn{1}{c}{var} & \multicolumn{1}{c}{$R^2$} \\ \cmidrule(l){2-9} 
Contemp & $0.71 \pm 0.16$         & $0.54 \pm 0.22$           & $0.72 \pm 0.12$         & $0.52 \pm 0.15$           & $0.71 \pm 0.08$         & $0.50 \pm 0.12$           & $0.72 \pm 0.06$         & $0.50 \pm 0.08$           \\
Lagged  & $0.56 \pm 0.18$         & $0.49 \pm 0.18$           & $0.54 \pm 0.13$         & $0.49 \pm 0.13$           & $0.54 \pm 0.11$         & $0.50 \pm 0.11$           & $0.54 \pm 0.10$         & $0.49 \pm 0.09$           \\
Overall & $0.58 \pm 0.09$         & $0.51 \pm 0.10$           & $0.56 \pm 0.06$         & $0.51 \pm 0.06$           & $0.54 \pm 0.03$         & $0.50 \pm 0.03$           & $0.54 \pm 0.02$         & $0.50 \pm 0.02$           \\ \bottomrule
\end{tabular}%
}

\end{table}
\clearpage
\section{Investigation on Influence of Different Degrees}\label{app:degree}
\subsection{On Varsortability}

\begin{figure}[h!]
    \centering
    \includegraphics[width = 0.6 \textwidth]{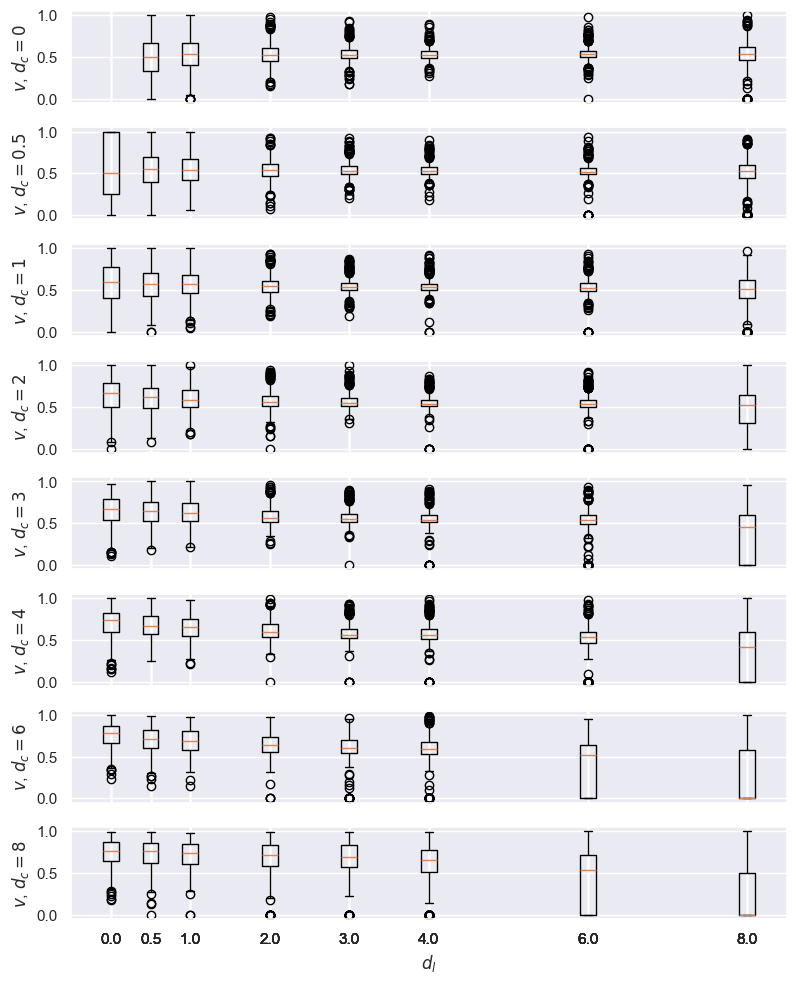}
    \caption{Varsortability for $d = 10$ nodes for different contemporaneous degrees $d_c$ and lagged degrees $d_l$ }
    \label{fig:comp_var_10}
\end{figure}

\begin{figure}[h!]
    \centering
    \includegraphics[width = 0.6 \textwidth]{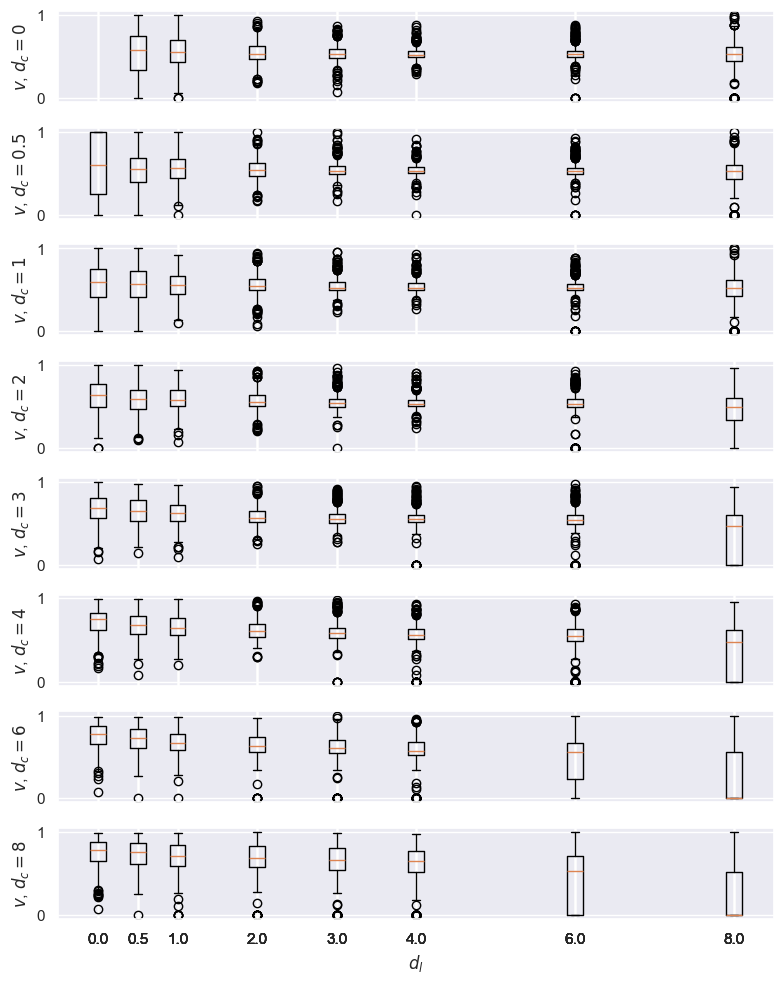}
    \caption{Varsortability for $d = 20$ nodes for different contemporaneous degrees $d_c$ and lagged degrees $d_l$ }
    \label{fig:comp_var_20}
\end{figure}

\clearpage

\subsection{On R2-sortability}
\begin{figure}[h!]
    \centering
    \includegraphics[width = 0.6 \textwidth]{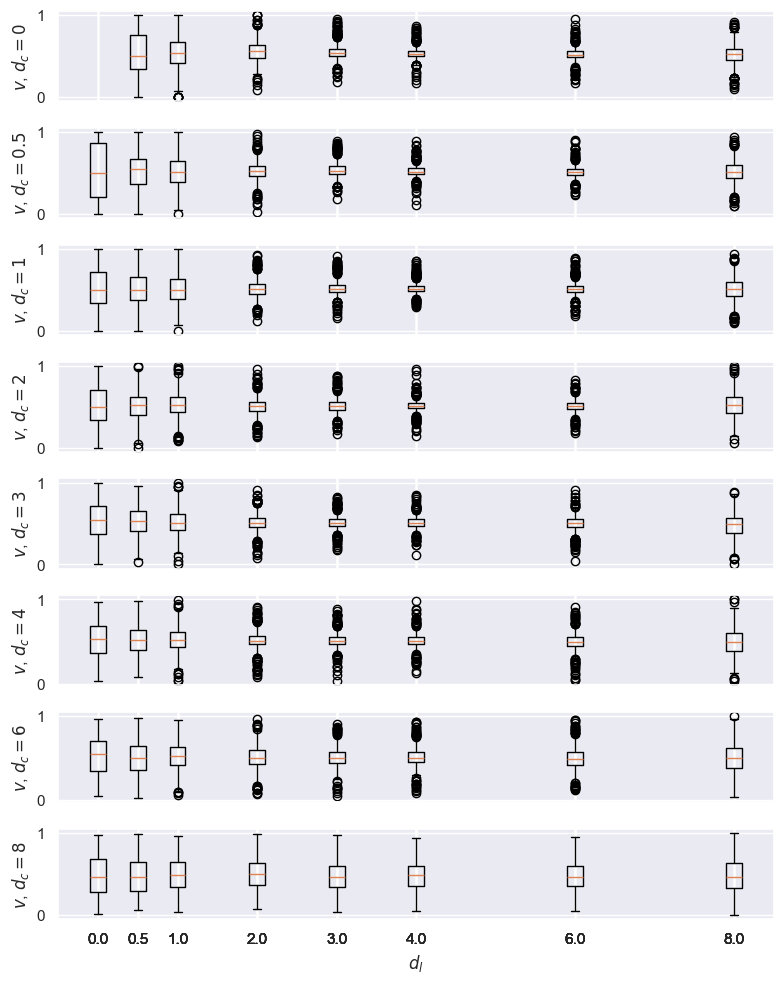}
    \caption{$R^2$-sortability for $d = 10$ nodes for different contemporaneous degrees $d_c$ and lagged degrees $d_l$}
    \label{fig:comp_r2_10}
\end{figure}

\begin{figure}[h!]
    \centering
    \includegraphics[width = 0.6 \textwidth]{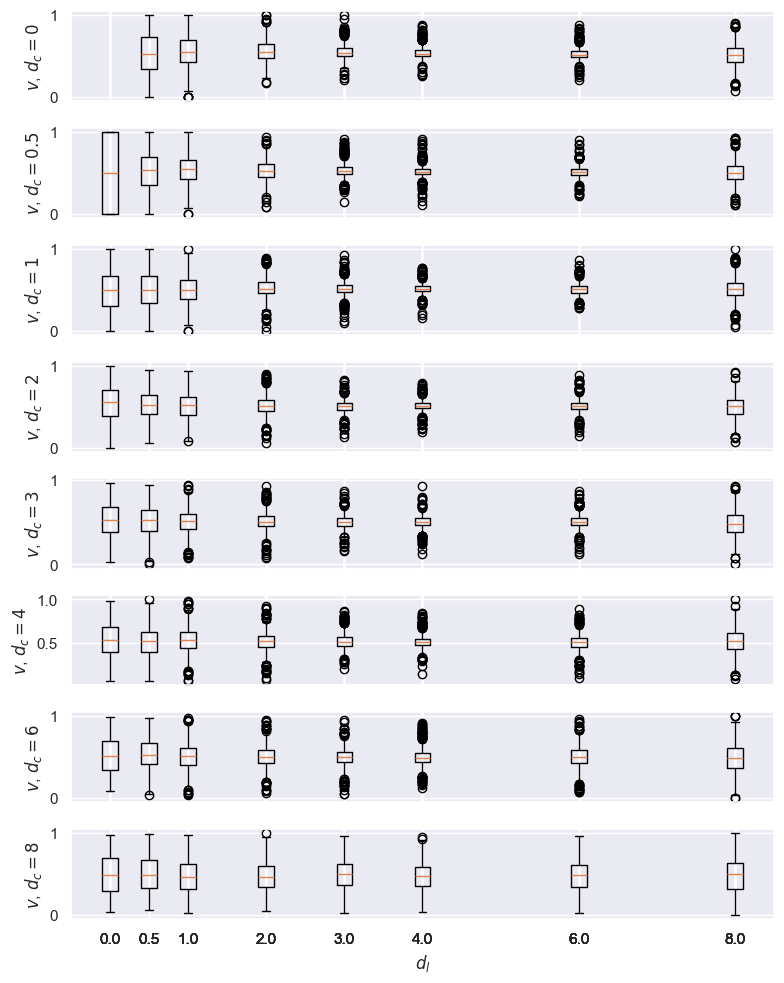}
    \caption{$R^2$-sortability for $d = 20$ nodes for different contemporaneous degrees $d_c$ and lagged degrees $d_l$} 
    \label{fig:comp_r2_20}
\end{figure}

\clearpage

\section{Results Causalchamber Data}\label{app:chamber}

\begin{longtable}{@{}llrr@{}}
\toprule
Dataset                         & Experiment                            & varsortability & $R^2$-sortability \\* \midrule
\endfirsthead
\endhead
\bottomrule
\endfoot
\endlastfoot
lt\_camera\_walks\_v1           & color\_mix                            & 0.95           & 0.28              \\
lt\_camera\_walks\_v1           & actuator\_mix                         & 0.95           & 0.18              \\
lt\_color\_regression\_v1       & pol\_1\_90                            & 0.89           & 0.21              \\
lt\_color\_regression\_v1       & aperture\_11.0                        & 0.95           & 0.16              \\
lt\_color\_regression\_v1       & iso\_1000.0                           & 0.95           & 0.18              \\
lt\_color\_regression\_v1       & bright\_colors                        & 0.95           & 0.00              \\
lt\_color\_regression\_v1       & shutter\_speed\_0.002                 & 0.95           & 0.19              \\
lt\_color\_regression\_v1       & aperture\_5.0                         & 0.95           & 0.16              \\
lt\_color\_regression\_v1       & pol\_1\_45                            & 0.95           & 0.16              \\
lt\_color\_regression\_v1       & reference                             & 0.95           & 0.18              \\
lt\_color\_regression\_v1       & iso\_500.0                            & 0.95           & 0.16              \\
lt\_color\_regression\_v1       & shutter\_speed\_0.001                 & 0.95           & 0.14              \\
lt\_interventions\_standard\_v1 & uniform\_t\_ir\_1\_mid                & 0.95           & 0.44              \\
lt\_interventions\_standard\_v1 & uniform\_diode\_vis\_1\_mid           & 0.95           & 0.47              \\
lt\_interventions\_standard\_v1 & uniform\_osr\_angle\_2\_mid           & 0.95           & 0.51              \\
lt\_interventions\_standard\_v1 & uniform\_l\_12\_mid                   & 0.95           & 0.49              \\
lt\_interventions\_standard\_v1 & uniform\_diode\_ir\_3\_mid            & 0.95           & 0.49              \\
lt\_interventions\_standard\_v1 & uniform\_t\_ir\_1\_strong             & 0.95           & 0.40              \\
lt\_interventions\_standard\_v1 & uniform\_diode\_ir\_3\_strong         & 0.95           & 0.47              \\
lt\_interventions\_standard\_v1 & uniform\_red\_mid                     & 0.95           & 0.47              \\
lt\_interventions\_standard\_v1 & uniform\_diode\_ir\_2\_mid            & 0.95           & 0.46              \\
lt\_interventions\_standard\_v1 & uniform\_osr\_angle\_2\_strong        & 0.95           & 0.47              \\
lt\_interventions\_standard\_v1 & uniform\_t\_vis\_3\_strong            & 0.82           & 0.46              \\
lt\_interventions\_standard\_v1 & uniform\_pol\_1\_strong               & 0.86           & 0.46              \\
lt\_interventions\_standard\_v1 & uniform\_osr\_angle\_2\_weak          & 0.95           & 0.46              \\
lt\_interventions\_standard\_v1 & uniform\_pol\_2\_mid                  & 0.95           & 0.46              \\
lt\_interventions\_standard\_v1 & uniform\_t\_ir\_3\_weak               & 0.95           & 0.49              \\
lt\_interventions\_standard\_v1 & uniform\_t\_ir\_2\_weak               & 0.95           & 0.47              \\
lt\_interventions\_standard\_v1 & uniform\_diode\_vis\_2\_mid           & 0.95           & 0.47              \\
lt\_interventions\_standard\_v1 & uniform\_t\_vis\_1\_weak              & 0.95           & 0.46              \\
lt\_interventions\_standard\_v1 & uniform\_l\_11\_mid                   & 0.95           & 0.53              \\
lt\_interventions\_standard\_v1 & uniform\_osr\_angle\_1\_mid           & 0.95           & 0.44              \\
lt\_interventions\_standard\_v1 & uniform\_v\_angle\_1\_strong          & 0.91           & 0.53              \\
lt\_interventions\_standard\_v1 & uniform\_osr\_c\_weak                 & 0.95           & 0.47              \\
lt\_interventions\_standard\_v1 & uniform\_t\_ir\_2\_mid                & 0.95           & 0.44              \\
lt\_interventions\_standard\_v1 & uniform\_osr\_c\_strong               & 0.95           & 0.42              \\
lt\_interventions\_standard\_v1 & uniform\_diode\_ir\_1\_strong         & 0.95           & 0.47              \\
lt\_interventions\_standard\_v1 & uniform\_t\_ir\_3\_mid                & 0.95           & 0.47              \\
lt\_interventions\_standard\_v1 & uniform\_t\_ir\_3\_strong             & 0.91           & 0.42              \\
lt\_interventions\_standard\_v1 & uniform\_pol\_1\_mid                  & 0.91           & 0.47              \\
lt\_interventions\_standard\_v1 & uniform\_diode\_vis\_3\_mid           & 0.86           & 0.44              \\
lt\_interventions\_standard\_v1 & uniform\_diode\_ir\_1\_mid            & 0.95           & 0.46              \\
lt\_interventions\_standard\_v1 & uniform\_blue\_mid                    & 0.95           & 0.37              \\
lt\_interventions\_standard\_v1 & uniform\_osr\_c\_mid                  & 0.95           & 0.47              \\
lt\_interventions\_standard\_v1 & uniform\_t\_vis\_1\_strong            & 0.95           & 0.47              \\
lt\_interventions\_standard\_v1 & uniform\_green\_mid                   & 0.95           & 0.49              \\
lt\_interventions\_standard\_v1 & uniform\_diode\_ir\_2\_strong         & 0.95           & 0.47              \\
lt\_interventions\_standard\_v1 & uniform\_osr\_angle\_1\_weak          & 0.95           & 0.46              \\
lt\_interventions\_standard\_v1 & uniform\_v\_angle\_2\_mid             & 0.95           & 0.53              \\
lt\_interventions\_standard\_v1 & uniform\_t\_ir\_1\_weak               & 0.95           & 0.46              \\
lt\_interventions\_standard\_v1 & uniform\_t\_vis\_2\_strong            & 0.91           & 0.46              \\
lt\_interventions\_standard\_v1 & uniform\_blue\_strong                 & 0.95           & 0.32              \\
lt\_interventions\_standard\_v1 & uniform\_t\_vis\_1\_mid               & 0.95           & 0.49              \\
lt\_interventions\_standard\_v1 & uniform\_l\_32\_mid                   & 0.95           & 0.47              \\
lt\_interventions\_standard\_v1 & uniform\_l\_22\_mid                   & 0.95           & 0.49              \\
lt\_interventions\_standard\_v1 & uniform\_v\_c\_mid                    & 0.95           & 0.47              \\
lt\_interventions\_standard\_v1 & uniform\_green\_strong                & 0.95           & 0.42              \\
lt\_interventions\_standard\_v1 & uniform\_v\_angle\_2\_strong          & 0.91           & 0.51              \\
lt\_interventions\_standard\_v1 & uniform\_t\_vis\_3\_mid               & 0.86           & 0.46              \\
lt\_interventions\_standard\_v1 & uniform\_osr\_angle\_1\_strong        & 0.95           & 0.47              \\
lt\_interventions\_standard\_v1 & uniform\_t\_ir\_2\_strong             & 0.95           & 0.42              \\
lt\_interventions\_standard\_v1 & uniform\_v\_angle\_1\_mid             & 0.95           & 0.46              \\
lt\_interventions\_standard\_v1 & uniform\_pol\_2\_strong               & 0.86           & 0.49              \\
lt\_interventions\_standard\_v1 & uniform\_red\_strong                  & 0.95           & 0.51              \\
lt\_interventions\_standard\_v1 & uniform\_v\_c\_strong                 & 1.00           & 0.46              \\
lt\_interventions\_standard\_v1 & uniform\_t\_vis\_2\_mid               & 0.95           & 0.42              \\
lt\_interventions\_standard\_v1 & uniform\_t\_vis\_2\_weak              & 0.95           & 0.42              \\
lt\_interventions\_standard\_v1 & uniform\_t\_vis\_3\_weak              & 0.91           & 0.47              \\
lt\_interventions\_standard\_v1 & uniform\_l\_31\_mid                   & 0.95           & 0.47              \\
lt\_interventions\_standard\_v1 & uniform\_l\_21\_mid                   & 0.95           & 0.49              \\
lt\_interventions\_standard\_v1 & uniform\_reference                    & 0.95           & 0.49              \\
lt\_walks\_v1                   & actuators\_white                      & 0.96           & 0.26              \\
lt\_walks\_v1                   & color\_mix                            & 0.84           & 0.21              \\
wt\_walks\_v1                   & actuators\_random\_walk\_9            & 0.93           & 0.26              \\
wt\_walks\_v1                   & actuators\_random\_walk\_8            & 0.93           & 0.19              \\
wt\_walks\_v1                   & loads\_hatch\_mix\_slow\_run\_2       & 0.93           & 0.24              \\
wt\_walks\_v1                   & actuators\_random\_walk\_6            & 0.93           & 0.36              \\
wt\_walks\_v1                   & actuators\_random\_walk\_7            & 0.88           & 0.24              \\
wt\_walks\_v1                   & loads\_hatch\_mix\_slow\_run\_3       & 1.00           & 0.36              \\
wt\_walks\_v1                   & loads\_hatch\_mix\_slow\_run\_1       & 0.93           & 0.29              \\
wt\_walks\_v1                   & actuators\_random\_walk\_5            & 0.95           & 0.29              \\
wt\_walks\_v1                   & actuators\_random\_walk\_4            & 0.90           & 0.21              \\
wt\_walks\_v1                   & loads\_hatch\_mix\_slow\_run\_4       & 0.93           & 0.29              \\
wt\_walks\_v1                   & actuators\_random\_walk\_1            & 0.86           & 0.14              \\
wt\_walks\_v1                   & loads\_hatch\_mix\_slow\_run\_5       & 0.93           & 0.26              \\
wt\_walks\_v1                   & actuators\_random\_walk\_3            & 0.95           & 0.24              \\
wt\_walks\_v1                   & actuators\_random\_walk\_2            & 0.88           & 0.24              \\
wt\_walks\_v1                   & actuators\_random\_walk\_11           & 0.93           & 0.31              \\
wt\_walks\_v1                   & actuators\_random\_walk\_10           & 0.86           & 0.21              \\
wt\_walks\_v1                   & actuators\_random\_walk\_12           & 0.93           & 0.31              \\
wt\_walks\_v1                   & loads\_hatch\_mix\_fast\_run\_5       & 0.93           & 0.33              \\
wt\_walks\_v1                   & loads\_hatch\_mix\_fast\_run\_4       & 0.93           & 0.33              \\
wt\_walks\_v1                   & actuators\_random\_walk\_13           & 0.90           & 0.29              \\
wt\_walks\_v1                   & loads\_hatch\_mix\_fast\_run\_1       & 0.93           & 0.33              \\
wt\_walks\_v1                   & actuators\_random\_walk\_16           & 0.88           & 0.24              \\
wt\_walks\_v1                   & actuators\_random\_walk\_14           & 0.93           & 0.24              \\
wt\_walks\_v1                   & loads\_hatch\_mix\_fast\_run\_3       & 0.93           & 0.31              \\
wt\_walks\_v1                   & loads\_hatch\_mix\_fast\_run\_2       & 1.00           & 0.36              \\
wt\_walks\_v1                   & actuators\_random\_walk\_15           & 0.93           & 0.31              \\
lt\_malus\_v1                   & red\_255                              & 1.00           & 0.04              \\
lt\_malus\_v1                   & white\_128                            & 1.00           & 0.02              \\
lt\_malus\_v1                   & green\_64                             & 0.96           & 0.02              \\
lt\_malus\_v1                   & blue\_255                             & 1.00           & 0.02              \\
lt\_malus\_v1                   & green\_128                            & 0.96           & 0.00              \\
lt\_malus\_v1                   & white\_64                             & 1.00           & 0.02              \\
lt\_malus\_v1                   & blue\_64                              & 0.96           & 0.02              \\
lt\_malus\_v1                   & green\_255                            & 1.00           & 0.00              \\
lt\_malus\_v1                   & blue\_128                             & 1.00           & 0.04              \\
lt\_malus\_v1                   & white\_255                            & 1.00           & 0.02              \\
lt\_malus\_v1                   & red\_64                               & 0.96           & 0.02              \\
lt\_malus\_v1                   & red\_128                              & 0.96           & 0.04              \\
wt\_bernoulli\_v1               & random\_loads\_both                   & 1.00           & 0.00              \\
wt\_bernoulli\_v1               & fans\_off                             & 0.90           & 0.10              \\
wt\_bernoulli\_v1               & random\_loads\_intake                 & 1.00           & 0.07              \\
wt\_changepoints\_v1            & load\_in\_seed\_8                     & 0.94           & 0.13              \\
wt\_changepoints\_v1            & load\_in\_seed\_9                     & 0.98           & 0.10              \\
wt\_changepoints\_v1            & load\_in\_seed\_4                     & 0.94           & 0.13              \\
wt\_changepoints\_v1            & load\_in\_seed\_5                     & 0.94           & 0.15              \\
wt\_changepoints\_v1            & load\_in\_seed\_7                     & 0.94           & 0.13              \\
wt\_changepoints\_v1            & load\_in\_seed\_6                     & 0.94           & 0.15              \\
wt\_changepoints\_v1            & load\_in\_seed\_2                     & 0.94           & 0.15              \\
wt\_changepoints\_v1            & load\_in\_seed\_3                     & 0.94           & 0.15              \\
wt\_changepoints\_v1            & load\_in\_seed\_1                     & 0.94           & 0.15              \\
wt\_changepoints\_v1            & load\_in\_seed\_0                     & 0.94           & 0.15              \\
wt\_intake\_impulse\_v1         & load\_out\_0.5\_osr\_downwind\_4      & 1.00           & 0.29              \\
wt\_intake\_impulse\_v1         & load\_out\_0.5\_osr\_downwind\_2      & 1.00           & 0.29              \\
wt\_intake\_impulse\_v1         & load\_out\_1\_osr\_downwind\_4        & 1.00           & 0.26              \\
wt\_intake\_impulse\_v1         & load\_out\_0.5\_osr\_downwind\_8      & 1.00           & 0.29              \\
wt\_intake\_impulse\_v1         & load\_out\_0.01\_osr\_downwind\_4     & 1.00           & 0.45              \\
wt\_pressure\_control\_v1       & hatch\_0                              & 0.92           & 0.36              \\
lt\_test\_v1                    & current\_sensor                       & 1.00           & 0.04              \\
lt\_test\_v1                    & angle\_sensors                        & 0.96           & 0.00              \\
lt\_test\_v1                    & analog\_calibration                   & 0.93           & 0.00              \\
lt\_test\_v1                    & ir\_sensors                           & 0.96           & 0.00              \\
wt\_test\_v1                    & zero\_load                            & 1.00           & 0.21              \\
wt\_test\_v1                    & mic\_effects                          & 0.93           & 0.24              \\
wt\_test\_v1                    & potis\_coarse                         & 0.83           & 0.12              \\
wt\_test\_v1                    & tach\_resolution                      & 1.00           & 0.12              \\
wt\_test\_v1                    & osr\_mic                              & 1.00           & 0.00              \\
wt\_test\_v1                    & no\_load                              & 1.00           & 0.10              \\
wt\_test\_v1                    & potis\_fine                           & 0.86           & 0.12              \\
wt\_test\_v1                    & osr\_barometers                       & 0.81           & 0.10              \\
wt\_test\_v1                    & analog\_calibration                   & 1.00           & 0.00              \\
wt\_test\_v1                    & steps                                 & 0.93           & 0.43              \\
lt\_camera\_test\_v1            & polarizer\_effect\_bright             & 0.95           & 0.00              \\
lt\_camera\_test\_v1            & pure\_colors\_bright                  & 0.92           & 0.50              \\
lt\_camera\_test\_v1            & polarizer\_effect\_dark               & 0.95           & 0.07              \\
lt\_camera\_test\_v1            & pure\_colors\_dark                    & 0.95           & 0.50              \\
lt\_camera\_test\_v1            & palette                               & 0.95           & 0.18              \\
wt\_validate\_v1                & validate\_v\_2                        & 1.00           & 0.00              \\
wt\_validate\_v1                & validate\_v\_in                       & 1.00           & 0.02              \\
wt\_validate\_v1                & validate\_load\_out\_pressure\_intake & 1.00           & 0.00              \\
wt\_validate\_v1                & validate\_v\_out                      & 1.00           & 0.00              \\
wt\_validate\_v1                & validate\_load\_out\_current\_in      & 1.00           & 0.02              \\
wt\_validate\_v1                & validate\_v\_1                        & 1.00           & 0.00              \\
wt\_validate\_v1                & validate\_osr\_1                      & 1.00           & 0.00              \\
wt\_validate\_v1                & validate\_pot\_1                      & 1.00           & 0.00              \\
wt\_validate\_v1                & validate\_osr\_downwind               & 0.88           & 0.12              \\
wt\_validate\_v1                & validate\_res\_out                    & 1.00           & 0.00              \\
wt\_validate\_v1                & validate\_osr\_2                      & 1.00           & 0.02              \\
wt\_validate\_v1                & validate\_pot\_2                      & 1.00           & 0.00              \\
wt\_validate\_v1                & validate\_load\_out\_mic              & 1.00           & 0.00              \\
wt\_validate\_v1                & validate\_load\_in\_mic               & 1.00           & 0.02              \\
wt\_validate\_v1                & validate\_load\_in\_current\_out      & 1.00           & 0.02              \\
wt\_validate\_v1                & validate\_hatch\_rpms                 & 0.90           & 0.02              \\
wt\_validate\_v1                & validate\_osr\_out                    & 1.00           & 0.02              \\
wt\_validate\_v1                & validate\_osr\_upwind                 & 0.88           & 0.12              \\
wt\_validate\_v1                & validate\_res\_in                     & 1.00           & 0.02              \\
wt\_validate\_v1                & validate\_load\_in                    & 1.00           & 0.00              \\
wt\_validate\_v1                & validate\_v\_mic                      & 1.00           & 0.00              \\
wt\_validate\_v1                & validate\_osr\_intake                 & 0.88           & 0.12              \\
wt\_validate\_v1                & validate\_load\_out                   & 1.00           & 0.02              \\
wt\_validate\_v1                & validate\_osr\_ambient                & 0.90           & 0.12              \\
wt\_validate\_v1                & validate\_osr\_in                     & 1.00           & 0.00              \\
wt\_validate\_v1                & validate\_osr\_mic                    & 1.00           & 0.00              \\
wt\_validate\_v1                & validate\_hatch\_mic                  & 0.93           & 0.07              \\
wt\_validate\_v1                & validate\_hatch\_pressures            & 0.90           & 0.02              \\
wt\_pc\_validate\_v1            & validate\_pressure\_downwind\_loads   & 0.78           & 0.14              \\
lt\_validate\_v1                & validate\_l\_11                       & 0.96           & 0.00              \\
lt\_validate\_v1                & validate\_osr\_c                      & 0.98           & 0.02              \\
lt\_validate\_v1                & validate\_l\_12                       & 0.96           & 0.02              \\
lt\_validate\_v1                & validate\_v\_c                        & 1.00           & 0.00              \\
lt\_validate\_v1                & validate\_osr\_angle\_2               & 0.98           & 0.02              \\
lt\_validate\_v1                & validate\_red                         & 0.96           & 0.09              \\
lt\_validate\_v1                & validate\_osr\_angle\_1               & 0.98           & 0.02              \\
lt\_validate\_v1                & validate\_pol\_1                      & 0.94           & 0.06              \\
lt\_validate\_v1                & validate\_diode\_vis\_2               & 1.00           & 0.02              \\
lt\_validate\_v1                & validate\_diode\_vis\_3               & 1.00           & 0.02              \\
lt\_validate\_v1                & validate\_pol\_2                      & 1.00           & 0.02              \\
lt\_validate\_v1                & validate\_diode\_vis\_1               & 1.00           & 0.02              \\
lt\_validate\_v1                & validate\_diode\_ir\_1                & 1.00           & 0.02              \\
lt\_validate\_v1                & validate\_v\_angle\_1                 & 1.00           & 0.02              \\
lt\_validate\_v1                & validate\_diode\_ir\_2                & 1.00           & 0.02              \\
lt\_validate\_v1                & validate\_diode\_ir\_3                & 1.00           & 0.02              \\
lt\_validate\_v1                & validate\_v\_angle\_2                 & 1.00           & 0.03              \\
lt\_validate\_v1                & validate\_green                       & 0.98           & 0.00              \\
lt\_validate\_v1                & validate\_blue                        & 0.98           & 0.00              \\
lt\_validate\_v1                & validate\_l\_31                       & 0.96           & 0.02              \\
lt\_validate\_v1                & validate\_l\_32                       & 0.96           & 0.00              \\
lt\_validate\_v1                & validate\_t\_ir\_1                    & 1.00           & 0.00              \\
lt\_validate\_v1                & validate\_t\_vis\_3                   & 1.00           & 0.00              \\
lt\_validate\_v1                & validate\_t\_vis\_2                   & 1.00           & 0.00              \\
lt\_validate\_v1                & validate\_l\_22                       & 0.96           & 0.02              \\
lt\_validate\_v1                & validate\_t\_ir\_2                    & 1.00           & 0.02              \\
lt\_validate\_v1                & validate\_t\_vis\_1                   & 1.00           & 0.02              \\
lt\_validate\_v1                & validate\_t\_ir\_3                    & 1.00           & 0.02              \\
lt\_validate\_v1                & validate\_l\_21                       & 0.98           & 0.02              \\
lt\_camera\_validate\_v1        & validate\_shutter\_speed              & 1.00           & 0.00              \\
lt\_camera\_validate\_v1        & validate\_iso                         & 1.00           & 0.00              \\
lt\_camera\_validate\_v1        & validate\_green                       & 0.98           & 0.00              \\
lt\_camera\_validate\_v1        & validate\_pol\_2                      & 1.00           & 0.00              \\
lt\_camera\_validate\_v1        & validate\_blue                        & 0.98           & 0.00              \\
lt\_camera\_validate\_v1        & validate\_aperture                    & 1.00           & 0.00              \\
lt\_camera\_validate\_v1        & validate\_red                         & 0.96           & 0.09              \\
lt\_camera\_validate\_v1        & validate\_pol\_1                      & 1.00           & 0.00              \\* \bottomrule
\end{longtable}



\end{document}

%% file: math_commands.tex
\usepackage{amsmath,amsfonts,bm}

\def\eqref#1{equation~\ref{#1}}

\def\1{\bm{1}}

\DeclareMathAlphabet{\mathsfit}{\encodingdefault}{\sfdefault}{m}{sl}
\SetMathAlphabet{\mathsfit}{bold}{\encodingdefault}{\sfdefault}{bx}{n}

